\documentclass{article}

\usepackage{listings}
\usepackage[T1]{fontenc}
\usepackage[export]{adjustbox}
\usepackage{lmodern}
\usepackage{booktabs}
\usepackage{multirow}
\usepackage{xspace}
\newcommand{\cabrnet}{CaBRNet\xspace}
\newcommand{\prototree}{ProtoTree\xspace}
\newcommand{\protopnet}{ProtoPNet\xspace}
\newcommand{\suggestedit}[1]{}
\newcommand{\yaml}{\texttt{YAML}\xspace}
\newcommand{\ie}{\emph{i.e., \xspace}}
\newcommand{\wrt}{\emph{{w.r.t.} \xspace }}
\newcommand{\eg}{\emph{e.g.,\xspace}\xspace}
\newcommand{\etc}{\emph{etc.}\xspace}
\usepackage{graphicx}
\usepackage{hyperref}

\begin{document}

\title{CaBRNet, An Open-Source Library For Developing And Evaluating Case-Based Reasoning Models}

\author{Romain Xu-Darme, Aymeric Varasse, Alban Grastien, \\Julien Girard-Satabin, Zakaria Chihani\\ \\
Université Paris-Saclay, CEA, List, F-91120, Palaiseau, France \\
{\tt\small romain.xu-darme(at)cea.fr}}
\date{May 15, 2024}


\maketitle
\begin{abstract}
  In the field of explainable AI, a vibrant effort is dedicated to the design of self-explainable models, as a more principled alternative to \textit{post-hoc} methods that attempt to explain the decisions after a model opaquely makes them. However, this productive line of research suffers from common downsides: lack of reproducibility, unfeasible comparison, diverging standards. In this paper, we propose \cabrnet, an open-source, modular, backward-compatible framework for \textbf{Ca}se-\textbf{B}ased \textbf{R}easoning \textbf{Net}works:  \href{https://github.com/aiser-team/cabrnet}{https://github.com/aiser-team/cabrnet}.
  \end{abstract}
\section{Introduction}
As a reflection of the social and ethical concerns related to the increasing use of AI-based systems in modern society, the field of explainable AI (XAI) has gained tremendous momentum in recent years. 
XAI mainly consists of two complementary avenues of research that aim at shedding some light into the inner-workings of complex ML models. 
On the one hand, \textit{post-hoc} explanation methods apply to existing models that have often been trained with the sole purpose of accomplishing a given task as efficiently as possible (\eg accuracy in a classification task).
On the other hand, self-explainable models are designed and trained to produce their own explanations along with their decision. 
The appeal of self-explainable models resides in the fact that rather than using an approximation (\ie a \textit{post-hoc} explanation method) to understand a complex model, it is better to directly enforce a simpler (and more understandable) decision-making process during the design and training of the ML model, provided that such a model would exhibit an acceptable level of performance.

In the case of image classification, works such as \protopnet~\cite{chen2019this} and subsequent improvements (\prototree~\cite{nauta2021neural}, ProtoPool~\cite{rymarczyk2021interpretable}, ProtoPShare~\cite{rymarczyk2021protopshare}, TesNet~\cite{wang2021interpretable}, PIP-Net~\cite{nauta2023pip}) have shown that self-explainable models can display accuracy results on par with more opaque state-of-the-art architectures. 
Such models are based on the principle of \textit{case-based reasoning} (CBR), in which new instances of a problem (the classification task) are solved using comparisons with an existing body of knowledge that takes the form of a database of prototypical representations of class instances.
In other words, the classification of an object is explained by the fact that it \textit{looks like}~\cite{chen2019this} another object from the training set for which the class is known.
CBR classifiers undeniably represent a stepping stone towards more understandable models. Yet, several recent works~\cite{nauta2020this,hoffmann2021this,gautam2022this,xudarme2023sanity,sacha2023interpretability} have questioned the interpretability of such models, proposing several evaluation metrics to go beyond the mere accuracy of the model and to estimate the quality of the explanations generated.
In practice however, a systematic comparison between different proposals for CBR models using these different metrics can be difficult to carry out in the absence of a unified framework that could gather the resources provided by their respective authors (\eg source codes made available on GitHub).

In this context, we propose \href{https://github.com/aiser-team/cabrnet}{\textit{\cabrnet}} (\textbf{Ca}se-\textbf{B}ased \textbf{R}easoning \textbf{Net}works), an open-source PyTorch library that proposes a generic approach to CBR models.
\cabrnet focuses on modularity, backward compatibility, and reproducibility, allowing us to easily implement existing works from the state of the art, propose new and innovative architectures, and compare them within a unified framework using multiple dedicated evaluation metrics.

\section{\cabrnet: One framework to rule them all...}
In a field of research as wide as XAI, where the proliferation of tools and methods is increasing, seeking a common framework is a natural temptation that often leads to yet another tool that evolves besides the rest.
To minimize this risk, \cabrnet is designed with three main objectives, essential for a lasting acceptance: supporting past state-of-the-art methods through backward compatibility, facilitating present developments by striving for modularity, and ensuring their reusability in future works through reproducibility.

\subsection{Modularity}

\begin{figure}[t!]
  \begin{center}
    \includegraphics[width=\textwidth]{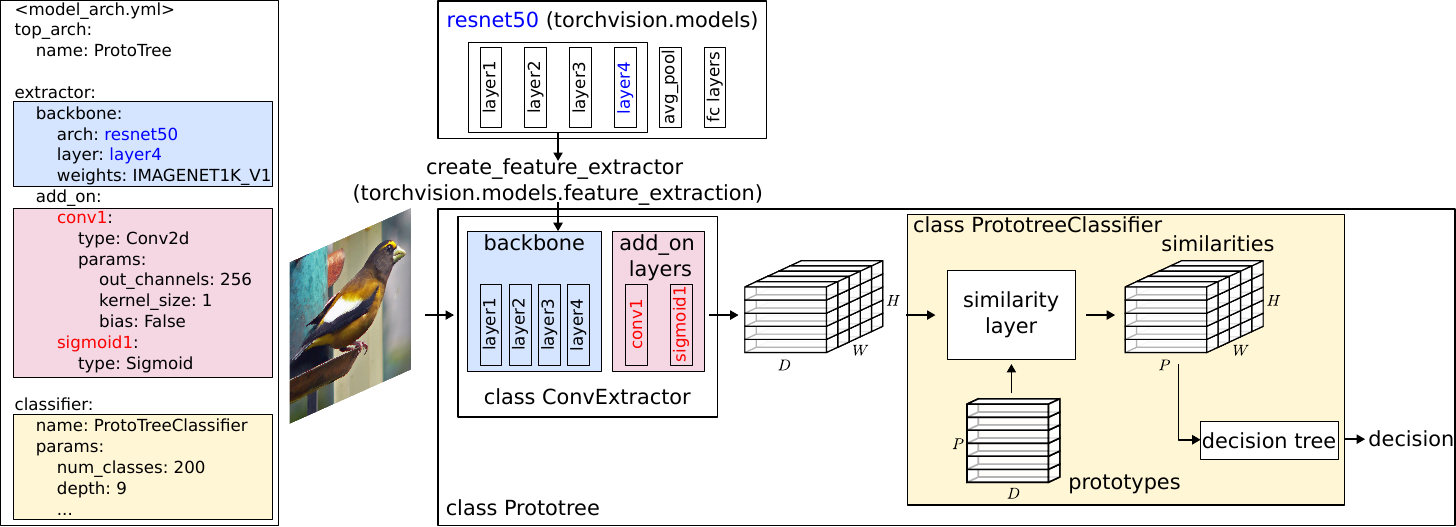}
  \end{center}
  \caption{Striving for modularity. From a YAML configuration file (left), \cabrnet backend exploits the common architecture of CBR image classifiers to simplify the instantiation of new models (here, a \prototree).}
  \label{fig::architecture}
\end{figure}

While each model has its own specific properties
(\eg decision tree in \prototree, scoring sheet in PIP-Net, linear layer in \protopnet), CBR image classifiers from the state of the art \cite{chen2019this,nauta2021neural,rymarczyk2021interpretable,rymarczyk2021protopshare,wang2021interpretable,nauta2023pip} share a common architecture,
displayed on Figure~\ref{fig::architecture}. In \cabrnet, \emph{this common architecture can be parametrized at will} using a set of \yaml files that are stored in each training directory for reproducibility (see Sec.~\ref{sec:reproducibility}).

\begin{table*}
  \caption{\cabrnet main configuration parameters (for a complete list, please refer to the \href{https://github.com/aiser-team/cabrnet/blob/master/docs/README.md}{documentation}).}\label{tab:params}
  \begin{tabular}{lll}
    \toprule
    \multicolumn{1}{c}{Parameter} & \multicolumn{1}{c}{Description} & \multicolumn{1}{c}{Supported values} \\
    \midrule
    & \multicolumn{1}{c}{Model configuration} & \\
    \midrule
    extractor/backbone/arch & Backbone of the extractor & Any from \href{https://pytorch.org/vision/stable/models.html}{torchvision.models} \\
    extractor/backbone/layer & Where to extract features & Any layer inside backbone \\
    extractor/add\_on & Add-on layers of extractor & Any from \href{https://pytorch.org/docs/stable/nn.html}{torch.nn} \\
    classifier & CBR classifier & \protopnet or \prototree \\
    \midrule
    & \multicolumn{1}{c}{Data configuration} \\
    \midrule
    train\_set/name & Training dataset class & Any from \href{https://pytorch.org/vision/main/datasets.html}{torchvision.datasets} \\
    train\_set/params/transform & Data preprocessing & Any from \href{https://pytorch.org/vision/stable/transforms.html}{torchvision.transforms}\\
    \midrule
    & \multicolumn{1}{c}{Visualization configuration} \\
    \midrule
    \multirow{2}{*}{attribution/type} & \multirow{2}{*}{Attribution method} & Upsampling, SmoothGrad, \\
    & & Backprop, PRP, RandGrads \\
    \multirow{1}{*}{view/type}& How to visualize patches& \\ & &\includegraphics[width=5cm]{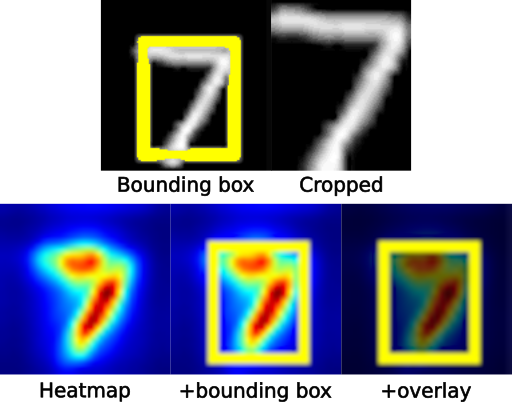}\\
    \bottomrule
  \end{tabular}
\end{table*}

More precisely, as shown in Figure~\ref{fig::architecture} and Table~\ref{tab:params}, the architecture of a CBR image classifier starts with a backbone -- usually the feature extractor of an existing convolutional neural network (CNN) such as ResNet50 -- with optional additional layers to reduce dimensionality
(\eg the fourth layer of a ResNet50 followed by a convolutional layer with sigmoid activation).

The feature extractor outputs a set of vectors, that form the latent space, and is followed by a similarity layer whose role is to compute the similarity between the latent vectors of the input image and a series of reference vectors, called \emph{prototypes}, each obtained from an image from the training set.
From the similarity scores, the process of locating the prototypes inside images (attribution) and visualizing the relevant pixels (\eg bounding box, cropping, heatmaps) is also parametrized by a configuration file, as described in Table~\ref{tab:params}. 
Finally, a decision layer, parametrized by an architecture-dependent python class (\eg decision tree, linear layer), assigns a score to each class based on the similarity scores, the highest being the decision (object classification).

In addition to these architectural choices, \cabrnet also allows specifying the parameters necessary for the training of the models (\eg objective functions, number of epochs, optimizer to use, which parts of the neural network can be updated and when, training dataset, preprocessing). For more information regarding the training and data configuration, we refer the reader to the \cabrnet \href{https://github.com/aiser-team/cabrnet/blob/master/docs/README.md}{documentation}.

In its current version (v0.2), and as shown in Table~\ref{tab:params}, \cabrnet implements two architectures (\protopnet and \prototree) and five attribution methods:
\begin{itemize}
  \item Upsampling of the similarity map with cubic interpolation, as in \cite{chen2019this,nauta2021neural,rymarczyk2021protopshare,rymarczyk2021interpretable,wang2021interpretable,nauta2023pip};
  \item SmoothGrads\cite{smilkov2017smoothgrads}/backpropagation\cite{simonyan2014deep}, as proposed in \cite{xudarme2023sanity};
  \item PRP\cite{gautam2022this}, a variant of LRP\cite{bach2015pixel} with a propagation rule for the similarity layer;
  \item RandGrads, a dummy attribution method returning random gradients (used as a baseline when comparing attribution methods in \cite{xudarme2023sanity}).
\end{itemize}
In the coming months, we plan to add support for more architectures (see Sec.~\ref{sec:future_work}).

\subsection{Backward compatibility}
Significant work has already been carried out on CBR classifiers. Therefore, ensuring that any result
obtained with previously existing codebases (\eg model training) can be reused in the \cabrnet framework
is paramount and has been one of our main and earliest priorities, in an effort to ensure backward compatibility.

First, \emph{any previously trained model can be loaded and imported within the \cabrnet framework}, \ie models trained using the original code proposed by the authors of \protopnet and \prototree can be used by our framework for other
purposes (\eg benchmarks) without having to retrain the model
from scratch.

Second, our implementation of existing architectures (currently, \protopnet and \prototree) supports two modes: 
i) default mode, where all operations have been reimplemented so that they are as
up-to-date as possible with the latest PyTorch versions; ii) compatibility mode, as a sanity check (\ie to make sure
that our implementation does not deviate from previous implementations), whose purpose is
to be accurate with previous works at the operation level\footnote{Backward compatibility was rigorously tested using unit tests covering all aspects of the process, from data loading to model training, to pruning and prototype projection.}.

\subsection{Reproducibility and transparency}\label{sec:reproducibility}
Reproducibility is key for the transparency of research and good software
engineering alike. 
However, replicating machine learning (ML) results can be particularly
tricky\footnote{See \href{https://reproducible.cs.princeton.edu/}{https://reproducible.cs.princeton.edu/}}. One reason for this is the inherent randomness at the core of ML programs (and the lack of documentation of settings related to that
randomness). 
The rapid update pace of common ML frameworks leads
to the regular deprecation of APIs and features, which may make a code impossible to run without
directly changing its source, or going through the rabbit hole of dependency
hell. As an example, the
mixed-precision training setting, available from PyTorch v1.10, uses different
types for CPU and GPU computations, which may lead to different results.
One last part is the under-specification of the
dataset constitution (curation process, classes imbalance, \etc) and preprocessing
pipelines (test/train splits, data transformations).

Reproducibility can cover many notions.
In~\cite{mcdermott2021reproducibility}, it is defined as the following:
``A ML study is reproducible if readers can fully replicate the
exact results reported in the paper.'' We strive to provide a similar definition:
``the same set of parameters will always yield the same results'', with
the following limitations:
\begin{itemize}
    \item reproducibility can be ensured only for a given hardware/software
        configuration. The software configuration is specified through the
        \lstinline{requirements.txt}. We are aware that stricter
        software environment specifications do exist (for instance, stateless
        build systems like Nix\footnote{\url{https://nixos.org/}}).
        Integrating such solutions into \cabrnet would be an interesting
        prospect, but also comes with an additional engineering cost to keep the
        library's ease of use. While not currently supported, we plan to save 
        information regarding the hardware configuration
        that led to a given result;
    \item hyperparameters that seem innocuous may influence the results. For
        instance, the size of the data batches - \textbf{even during testing} -
        has a small influence on the results. Consequently, we also save this information
        inside configuration files.
\end{itemize}

To address the randomness variation, we initialize the various random number generators (RNGs)
using a fixed seed that is stored along the training parameters. Thus, while
loading a model's parameters, the seed used for its training will be loaded as
well, guaranteeing that given the same hardware and hyperparameter
configuration, the variability induced by the pseudo-randomness is identical.
We also support the possibility to save checkpoints at various steps of the process, and we include
the current state of all RNGs to restore the training process exactly as it was.

As we believe that reproducibility is improved by a good documentation, we also
provide detailed installation instructions, a tutorial, the API reference and
the \cabrnet backend reference. The major part of this
documentation is automatically generated from the source code, ensuring consistency
between the code and the documentation at all times.

\section{... and in the light evaluate them}
In this section, we describe our design choices for the \cabrnet benchmark, with the purpose of proposing a framework for the systematic evaluation of CBR models. 

\subsection{Going beyond accuracy}
\begin{figure}
  \centering
  \includegraphics[width=0.6\textwidth]{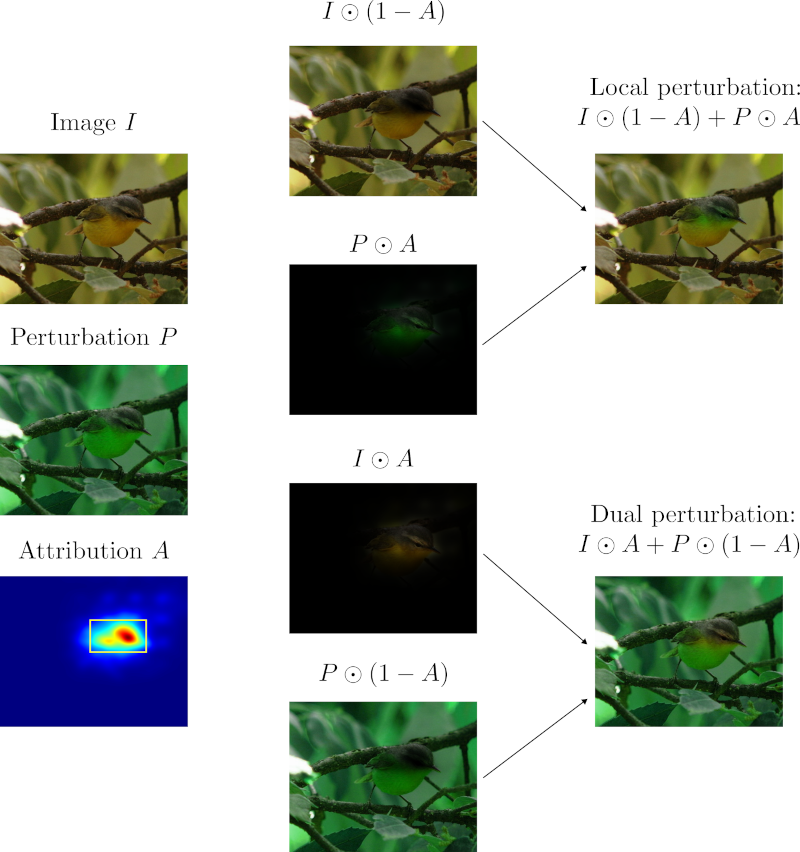}
  \caption{Improvement over the perturbation metric used in \cite{nauta2020this}. Rather than studying the drop in similarity score following a global perturbation of the image (\eg shift in the hue of the image), we apply a \emph{local} perturbation using the heatmap produced by the chosen attribution technique. Hence, we not only measure the sensitivity of the similarity score to a given perturbation, but also the ability of the attribution method to locate the most relevant pixels. Additionally, we measure the drop in similarity when applying the perturbation to \emph{anything but} the most important pixels (dual perturbation).}\label{fig:perturbation}
\end{figure}

Current CBR models rely on two assumptions: i) proximity in the latent space (\wrt a given distance metric) is equivalent to similarity in the visual space; ii) there exists a simple mapping between a latent vector and a localized region in the original image, due to the architecture of the feature extractor (CNN).
These assumptions have recently been put to the test by various metrics\footnote{Regarding the relevance and/or effectiveness of these metrics, we refer the reader to the original papers.}, that are already implemented or that we aim to implement in \cabrnet to streamline the evaluation of CBR models. In particular, \cabrnet currently integrates:
\begin{itemize}
  \item the perturbation-based explanation of \cite{nauta2020this}, with improvements: perturbations are no longer applied to the entire image, but to the identified patch of image corresponding to the prototypical part, as shown in Figure~\ref{fig:perturbation};
  \item the pointing game (relevance metric) of \cite{xudarme2023sanity}, with improvements: the metric now also supports the energy-based pointing game introduced in \cite{wang2019score}.
\end{itemize}
In the coming months, we plan to add support for more evaluation metrics (see Sec.~\ref{sec:future_work}).

\subsection{Stop wasting time retraining state-of-the-art models}
In our opinion, reproducing results from the state of the art (\ie re-training models) -- to serve as points of comparison with a new approach -- can waste valuable time and computing resources from AI researchers, when that time could be dedicated to improving that new approach. 
This issue mostly arises when proposing new evaluation metrics\footnote{When using common metrics (\eg model accuracy), it is possible to avoid retraining models by simply referring to the results provided by the original authors.} that must be applied to a wide range of state-of-the-art models. 
Thus, one of the objectives for \cabrnet is to \emph{publish pre-trained state-of-the-art CBR models, so that they can be used readily by the XAI research community}. 
For the sake of transparency, and as stated in Sec.~\ref{sec:reproducibility}, each published model is also associated with information ensuring a level of reproducibility, should a researcher wish to re-train these models. 
Moreover, in an effort to improve the statistical significance of all related experiments, \emph{we plan to publish at least three models per model configuration and dataset}. 
As an example, we have already published 6 models\footnote{Available at \href{https://zenodo.org/records/10894996}{https://zenodo.org/records/10894996}.} trained on the Caltech-UCSD Birds 200 (CUB200~\cite{welinder2010caltech}) using our implementation of \prototree with trees with depth 9 and 10.

\section{Conclusion and future works}\label{sec:future_work}
\cabrnet is open-source and welcomes any external contribution. We also strongly encourage the research community to publish trained models that can be reused within our evaluation framework. On our side, future developments include:
i) support for ProtoPool\cite{rymarczyk2021interpretable}, ProtoPShare\cite{rymarczyk2021protopshare}, PIP-Net\cite{nauta2023pip} and TesNet\cite{wang2021interpretable};
ii) support for the metric measuring the stability to JPEG compression \cite{hoffmann2021this} and the stability to adversarial attacks\cite{sacha2023interpretability};
iii) hands-on tutorials for gently introducing users to the various aspects of the framework.

Ideally, the modularity of \cabrnet design will appeal to both AI
researchers wanting to experiment with CBR approaches, and industrial actors
interested in deploying those approaches in a principled and traceable way.
Our goal is for \cabrnet to become the development framework for new and innovative CBR approaches by the community.


\newpage

\end{document}